\documentclass{article}
\usepackage{arxiv}
\usepackage[hidelinks]{hyperref}       

\usepackage{color}
\usepackage[style=numeric,sorting=none,natbib=true]{biblatex}
\usepackage{geometry}
\usepackage{amssymb}
\usepackage{latexsym}
\DeclareBibliographyCategory{nobibliography}
\addtocategory{nobibliography}{amos2022data}
\addtocategory{nobibliography}{ts2023data}
\addtocategory{nobibliography}{mmwhs2019data}
\addtocategory{nobibliography}{btcv2017data}
\addtocategory{nobibliography}{myo2024data}

\usepackage{url}
\usepackage{xcolor}

\usepackage{hyperref}
\usepackage{amsmath,amssymb,amsfonts}
\usepackage[T1]{fontenc}
\usepackage{microtype}
\usepackage{algorithmic}
\usepackage{graphicx}
\usepackage{calc}
\usepackage{booktabs}
\usepackage[bb=boondox]{mathalfa} 
\usepackage{siunitx} 
\usepackage{multirow} 
\newrobustcmd\B{\bfseries}

\usepackage{arydshln}

\usepackage{array}
\newcolumntype{P}[1]{>{\centering\arraybackslash}p{#1}}

\usepackage{textcomp}
\def\BibTeX{{\rm B\kern-.05em{\sc i\kern-.025em b}\kern-.08em
    T\kern-.1667em\lower.7ex\hbox{E}\kern-.125emX}}

\usepackage{subcaption}

\newcommand{\legendbox}[1]{%
    \textcolor{#1}{\rule{\fontcharht\font`X}{\fontcharht\font`X}}%
}

\usepackage{makecell}

\DeclareSIUnit\vox{vox}
\newcommand{\sipct}[1]{\SI[mode=text, retain-explicit-plus]{#1}{\percent}}
\DeclareSIUnit{\nothing}{\relax}
\newcommand{\sinum}[1]{\SI[mode=text, retain-explicit-plus]{#1}{\nothing}}
\newcommand{\simm}[1]{\SI[mode=text]{#1}{\milli\meter}}
\newcommand{\simmtuple}[3]{\sinum{#1}$\times$\sinum{#2}$\times$\SI[mode=text]{#3}{\cubic\milli\meter}}
\newcommand{\sivoxtuple}[3]{\sinum{#1}$\times$\sinum{#2}$\times$\SI[mode=text]{#3}{\vox}}
\newcommand{\simmpervoxtuple}[3]{\sinum{#1}$\times$\sinum{#2}$\times$\SI[mode=text, per-mode=symbol]{#3}{\cubic\milli\meter\per\vox}}

\newcommand{\fromto}[2]{\mbox{#1\,\textgreater\,#2}}
\newcommand{\bplotdescription}{%
    Ordinate shows Dice scores in \%. Median (---) and mean (+) are indicated for boxes. The significance of improvement over the source NNUNET BS base model is shown above boxes (\textsuperscript{*}\mbox{p\textless0.05}; \textsuperscript{**}\mbox{p\textless0.01}; \textsuperscript{***}\mbox{p\textless0.001}).
}

\definecolor{hi_1}{rgb}{0.9058823529411765, 0.2784313725490196, 0.3686274509803922}
\definecolor{hi_2}{rgb}{0.9411764705882353, 0.8470588235294118, 0.4745098039215686}
\definecolor{hi_3}{rgb}{0.4745098039215686, 0.8627450980392157, 0.9411764705882353}
\definecolor{hi_4}{rgb}{0.1411764705882353, 0.5333333333333333, 0.5333333333333333}

\definecolor{map10_1}{rgb}{0.90588, 0.27843, 0.36862}
\definecolor{map10_2}{rgb}{0.91750, 0.46574, 0.40350}
\definecolor{map10_3}{rgb}{0.92913, 0.65305, 0.43838}
\definecolor{map10_4}{rgb}{0.94117, 0.84705, 0.47450}
\definecolor{map10_5}{rgb}{0.78745, 0.85222, 0.62823}
\definecolor{map10_6}{rgb}{0.62823, 0.85757, 0.78745}
\definecolor{map10_7}{rgb}{0.47450, 0.86274, 0.94117}
\definecolor{map10_8}{rgb}{0.36078, 0.75035, 0.80202}
\definecolor{map10_9}{rgb}{0.25098, 0.64184, 0.66768}
\definecolor{map10_10}{rgb}{0.14117, 0.53333, 0.53333}

\definecolor{map5_1}{rgb}{0.90588, 0.27843, 0.36862}
\definecolor{map5_2}{rgb}{0.93245, 0.70657, 0.44835}
\definecolor{map5_3}{rgb}{0.70509, 0.85499, 0.71058}
\definecolor{map5_4}{rgb}{0.38823, 0.77748, 0.83561}
\definecolor{map5_5}{rgb}{0.14117, 0.53333, 0.53333}

\makeatletter
\if@twocolumn
\newcommand{\columnfigure}[5][]{
    \begin{figure}[#1]
        \centering
        \includegraphics[width=#3\columnwidth]{#2}
        \caption{#4}
        \label{#5}
    \end{figure}
}
\else
\newcommand{\columnfigure}[5][]{
    \begin{figure}{#1}
        \centering
        \includegraphics[width=\linewidth*\real{.5}*\real{#3}]{#2}
        \caption{#4}
        \label{#5}
    \end{figure}
}
\fi
\makeatother

\newcommand*{\dgttaPath}{./}
\newboolean{RENDER_TEXT_ONLY}
\setboolean{RENDER_TEXT_ONLY}{false}
\hypersetup{
    pdftitle={DG-TTA: Out-of-domain Medical Image Segmentation through Augmentation and Descriptor-driven Domain Generalization and Test-Time Adaptation},
    pdfsubject={DG-TTA},
    pdfauthor={Christian Weihsbach, Christian N.~Kruse, Alexander Bigalke, Mattias P.~Heinrich},
    pdfkeywords={Domain generalization, Domain-invariant descriptors, Test-time adaptation},
}

\title{DG-TTA: Out-of-domain Medical Image Segmentation through Augmentation and Descriptor-driven Domain Generalization and Test-Time Adaptation
}

\author{ 
    \href{https://orcid.org/0000-0002-3839-096X}{\includegraphics[scale=0.06]{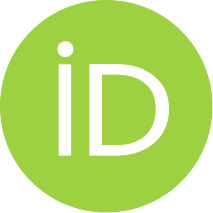}\hspace{1mm}Christian Weihsbach} \\
    Institute of Medical Informatics\\
    University of Lübeck\\
    23538 Lübeck, Germany \\
    \texttt{ch.weihsbach@student.uni-luebeck.de}\\
    \And
    \href{https://orcid.org/0000-0002-2322-4766}{\includegraphics[scale=0.06]{orcid.pdf}\hspace{1mm}Christian N.~Kruse} \\
    Dampsoft GmbH\\
    24351 Damp, Germany\\
    \texttt{christian.kruse@dampsoft.de}\\
    \And
    \href{https://orcid.org/0000-0001-7824-5735}{\includegraphics[scale=0.06]{orcid.pdf}\hspace{1mm}Alexander Bigalke} \\
    Institute of Medical Informatics\\
    University of Lübeck\\
    23538 Lübeck, Germany\\
    \texttt{alexander.bigalke@uni-luebeck.de}\\
    \And
    \href{https://orcid.org/0000-0002-7489-1972}{\includegraphics[scale=0.06]{orcid.pdf}\hspace{1mm}Mattias P.~Heinrich} \\
    Institute of Medical Informatics\\
    University of Lübeck\\
    23538 Lübeck, Germany \\
    \texttt{mattias.heinrich@uni-luebeck.de}\\
}

\renewcommand{\shorttitle}{DG-TTA}

\begin{document}
    \maketitle

    \begin{abstract}
        Purpose: Applying pre-trained medical deep learning segmentation models on out-of-domain images often yields predictions of insufficient quality. In this study, we propose to use a powerful generalizing descriptor along with augmentation to enable domain-generalized pre-training and test-time adaptation, achieving high-quality segmentation in unseen domains.

Materials and Methods: In this retrospective study five different publicly available datasets (2012 to 2022) including 3D CT and MRI images are used to evaluate segmentation performance in out-of-domain scenarios. The settings include abdominal, spine, and cardiac imaging. The data is randomly split into training and test samples. Domain-generalized pre-training on source data is used to obtain the best initial performance in the target domain. We introduce the combination of the generalizing SSC descriptor and GIN intensity augmentation for optimal generalization. Segmentation results are subsequently optimized at test time, where we propose to adapt the pre-trained models for every unseen scan with a consistency scheme using the same augmentation-descriptor combination. The segmentation is evaluated using Dice similarity and Hausdorff distance and the significance of improvements is tested with the Wilcoxon signed-rank test.

Results: The proposed generalized pre-training and subsequent test-time adaptation improves model performance significantly in CT to MRI cross-domain prediction for abdominal (\sipct{+46.2} and \sipct{+28.2} Dice), spine (\sipct{+72.9}), and cardiac (\sipct{+14.2} and \sipct{+55.7} Dice) scenarios (p\textless0.001).

Conclusion: Our method enables optimal, independent usage of medical image source and target data and bridges domain gaps successfully with a compact and efficient methodology.
Open-source code available at: \url{https://github.com/multimodallearning/DG-TTA}
    \end{abstract}

    \newcommand*{\nnunetBaseVal}{\sipct{32.0}}
\newcommand*{\targetDrop}{\sipct{-52.6}}
\newcommand*{\tentGain}{\sipct{+17.6}}
\newcommand*{\ttaRmiGain}{\sipct{+7.8}}
\newcommand*{\rsaAfterAdap}{\sipct{10.7}}
\newcommand*{\rsaBeforeAdaptation}{\sipct{2.7}}
\newcommand*{\adamiafteradaptation}{\sipct{62.8}}
\newcommand*{\ginBcvAmosVal}{\sipct{76.3}}
\newcommand*{\sscBtcvAmosVal}{\sipct{76.1}}
\newcommand*{\ginSscBtcvAmosVal}{\sipct{78.6}}
\newcommand*{\ginBtcvAmosAdapGain}{\sipct{+1.1}}
\newcommand*{\sscBtcvAmosAdapGain}{\sipct{+1.6}}
\newcommand*{\ginSscBtcvAmosAdapGain}{\sipct{-0.4}}
\newcommand*{\nnunetAdapTsAmosVal}{\sipct{64.1}}
\newcommand*{\nnunetBnAdapTsAmosVal}{\sipct{68.4}}
\newcommand*{\ginAdapTsAmosVal}{\sipct{81.4}}
\newcommand*{\sscAdapTsAmosVal}{\sipct{79.0}}
\newcommand*{\ginSscAdapTsAmosVal}{\sipct{79.6}}
\newcommand*{\ginSscTsSpineAdap}{\sipct{73.7}}
\newcommand*{\ginSscMmwhsCtMrAdapVal}{\sipct{71.5}}
\newcommand*{\ginSscVsGinOnlyGainMmwhsCtMr}{\sipct{+21.8}}
\newcommand*{\nnunetBnInternalGainBtcvAmosDice}{\sipct{+64.0}}
\newcommand*{\nnunetBnInternalReductionBtcvAmosHd}{\simm{-42.9}}
\newcommand*{\ginTsMmwhsMRVal}{\sipct{82.6}}
\newcommand*{\ginTsAmosAdapGain}{\sipct{+0.2}}
\newcommand*{\ginTsSpineAdapGain}{\sipct{-0.4}}
\newcommand*{\ginTsMmwhsMrAdapGain}{\sipct{+1.2}}
\newcommand*{\ginBtcvAmosHdVal}{\simm{27.9}}
\newcommand*{\ginSscAdapBtcvAmosHdVal}{\simm{17.2}}
\newcommand*{\ginSscAdapRank}{1.9}
\newcommand*{\sscAdapRank}{3.3}
\newcommand*{\ginAdapRank}{3.5}

\section{Introduction}
\label{sec:introduction}

    Medical image analysis, particularly image segmentation, has made a significant leap forward in recent years with deep learning. However, changes in data distribution introduced by different input modalities or devices can lead to errors in the performance of deep learning models \citep{karani2018lifelong}. Since multiple imaging techniques are often required for disease identification, treatment planning, and MRI devices especially offer broad flexibility in adjusting acquisition parameters, access to all imaging domains is usually infeasible. Consequently, trained models may produce inaccurate results when encountering unseen, out-of-domain data at test time \citep{pooch2020can}.

    Supervised finetuning can be used as a workaround to retrain or finetune networks for the unseen domain. Still, it would, in turn, require curating and labeling data again, which is often costly and time-consuming. Frequently studied approaches to overcome this effort use domain translation and unsupervised domain adaptation methods and often incorporate source images \citep{zhu2017unpaired,varsavsky2020test}.
    Accessing the source and target data jointly imposes a challenge since source data can be unavailable. For these reasons, source-free domain adaptation accesses source and target data separately in successive steps. Here, some methods require retraining on a larger set of target images to adapt models \citep{chen2021source,wen2023source}. In practice, a single out-of-domain data sample is often  given for which we want to obtain optimal results immediately. We target this setting in our study, facing the most challenging data constraints. In such a limited case domain generalization techniques can be used to optimize the model performance for `any' unseen out-of-distribution sample \citep{liu2023clip,xu2020robust,ouyang2022causality,hu2022domain,tobin2017domain,billot2023synthseg,bucci2021self,zhou2021models,he2022masked,hoyer2023mic}.
    Domain-generalization is an ultimate goal to achieve, but up to now, no universal solution that robustly works has been found.
    Test-time adaptation (TTA), as a complementary approach, optimizes the model performance only for one or a limited number of samples \citep{wang2020tent,sun2020test,bateson2020source,huang2022online,karani2021test,he2020self,he2021autoencoder,liu2022single}.

    We argue that linking both approaches enables optimal separate use of source and target data where domain generalization maximizes the base performance and TTA can further optimize the result.
    Numerous methods to bridge domain gaps have already been developed, but often require complex strategies and assumptions such as intertwined adaptation layers \citep{he2020self, he2021autoencoder}, indirect supervision tasks \citep{li2022self,huang2022online}, prior knowledge about label distributions \citep{bateson2020source}, assumptions on the distinctiveness of domains \citep{varsavsky2020test} or many consecutive steps \citep{zeng2024reliable}.

    We propose to employ DG-TTA, a minimally invasive and compact approach that uses a powerful augmentation-descriptor scheme  during domain-generalized pre-training and TTA for high-performance medical image segmentation in unseen domains under large domain gaps.

    \section{Materials and methods}
        \subsection{Study design and patients}
            We included data from five publicly available datasets in this retrospective study (see Tab. \ref{tab:dataset_characteristics_dgtta}).
            All patients included in the dataset studies have thus been previously reported \citep{landman2015miccai,ji2022amos,wasserthal2023totalsegmentator,burian2019lumbar,zhuang2019evaluation}.
            Those prior studies dealt with data collection and the development of individual segmentation methods whereas we target to develop a universal method for segmentation in this study. Throughout the next paragraphs, we use abbreviated names of the BTCV, AMOS, TotalSegmentator (TS), MyoSegmenTUM spine (SPINE), and MMWHS datasets. From the mentioned datasets cross-domain prediction settings are compiled, all targetting the difficult domain gap of CT source to MR target prediction (\fromto{CT}{MR}). Data selection was performed randomly and kept throughout all evaluated methods for fair comparison.
            We resampled all data samples to a uniform voxel size of \simmpervoxtuple{1.50}{1.50}{1.50}. For the SPINE task, we omit the spinous processes in the cross-domain \fromto{TS}{SPINE} prediction setting to provide comparable annotations.

            \begin{table}
                \caption{Characteristics of study patients of the publicly available datasets used in this study \citep{landman2015miccai,ji2022amos,wasserthal2023totalsegmentator,burian2019lumbar,zhuang2019evaluation}.}
                \label{tab:dataset_characteristics_dgtta}
                \resizebox{\textwidth}{!}{%
                    \begin{tabular}{p{0.23\linewidth}P{0.23\linewidth}P{0.23\linewidth}P{0.23\linewidth}P{0.23\linewidth}P{0.23\linewidth}}
                        \toprule
                        \textbf{Dataset} & \textbf{BTCV} & \textbf{AMOS} & \textbf{Total Segmentator Training dataset} & \textbf{MyoSegmenTUM spine} & \textbf{MMWHS} \\\midrule
                        Variable & & & & & \\
                        Date range & $\le$ 2015 & 2022 & 2012 --- 2020 & $\le$ 2018 & $\le$ 2017 \\
                        Modalities & CT& CT/MR & CT& MR & CT/MR \\
                        CT scans& 50 & 500& 1204 & 0& 60 \\
                        MRI scans & 0 & 100& 0 & 54 & 60 \\
                        Patients & 50& 600& 1204 & 54 & 60 \\
                        Sites & N/A& 1& 8 & 1& 3  \\
                        Scanners & N/A& 8& 16& 1& 4  \\
                        Sex  & & & & & \\
                        Male & N/A & 314 & $\sim$700 & 15 & N/A \\
                        Female & N/A& 186& $\sim$500 & 39 & N/A \\
                        Not reported & 50 & 0 & N/A & 0 & N/A \\
                        & & & & & \\
                        Age (y) & & & & & \\
                        Min  & N/A& 22 & 18& 21 & N/A \\
                        Max  & N/A& 85 & 100& 78 & N/A \\
                        Median & N/A& 50 & $\sim$70  & 40 & N/A \\
                        Mean & N/A& 48.7& $\sim$70.0& 51.6 & N/A \\
                        Labelled structures &
                        Spleen, right kidney, left kidney,   gallbladder, esophagus, liver, stomach, aorta, inferior vena cava, portal   vein and splenic vein, pancreas, right adrenal gland, left adrenal gland &
                        Spleen, right kidney, left kidney gallbladder, esophagus, liver, stomach, aorta, inferior vena cava, pancreas, right adrenal gland, left adrenal gland, duodenum, bladder, prostate / uterus &
                        used labelled structures: cardiac, abdominal organ and lumbar spine labels (a subset of the 27 organs, 59 bones, 10 muscles, and eight vessels labelled) &
                        Vertebral bodies L1 to L5 &
                        Myocardium of left ventricle, left ventricle, right ventricle, left atrium, right atrium, aortic trunk,   pulmonary artery trunk \\
                        Key clinical characteristics &
                        Patients were randomly selected from a combination of an ongoing colorectal cancer chemotherapy trial, and a retrospective ventral hernia study &
                        Patients were be diagnosed with abdominal tumors and abnormalities &
                        Patients with no signs of abnormality (404), patients with different types of abnormality (645), including tumor, vascular, trauma, inflammation, bleeding, and other &
                        Healthy volunteers &
                        Patients with pathologies involving cardiac diseases,  myocardium infarction, atrial fibrillation, tricuspid regurgitation, aortic valve stenosis, Alagille syndrome, Williams syndrome, dilated cardiomyopathy, aortic coarctation and Tetralogy of Fallot. \\\bottomrule
                    \end{tabular}
                }
            \end{table}
            \begin{figure*}
                \centerline{\includegraphics[width=\linewidth]{\dgttaPath/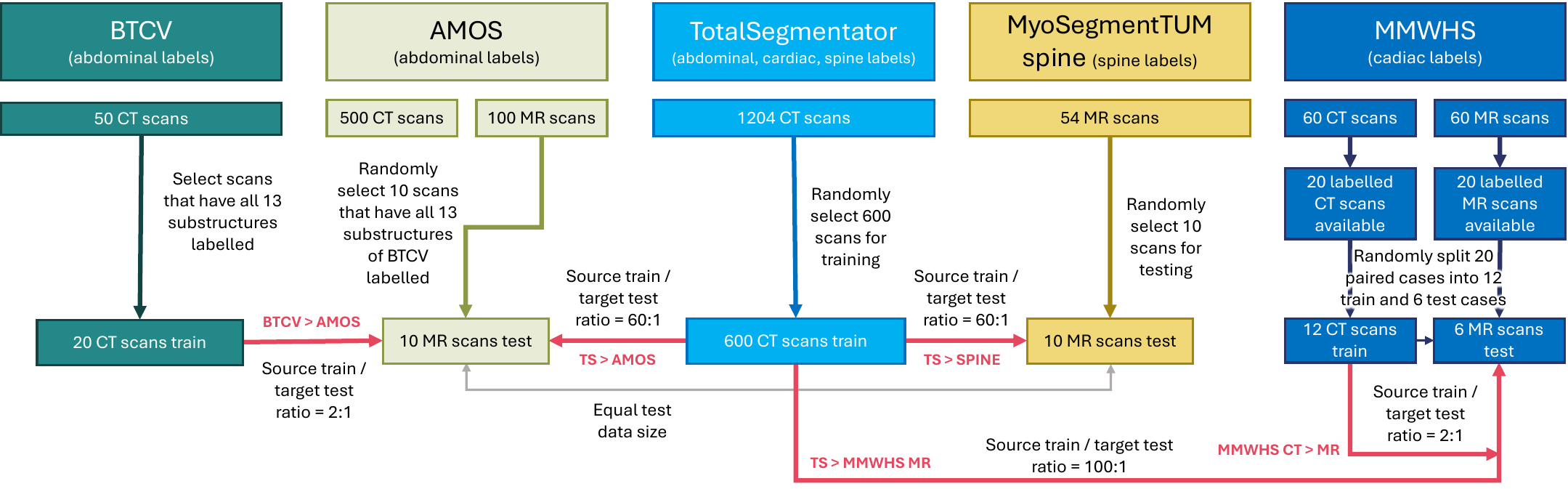}}
                \caption{Study flowchart. Data from five publicly available datasets was included and combined to result in several out-of-domain \fromto{CT}{MR} prediction scenarios (their combination is indicated by the red arrows)\citep{landman2015miccai,ji2022amos,wasserthal2023totalsegmentator,burian2019lumbar,zhuang2019evaluation}. We randomly extracted subsamples for a source and target data ratio of at least 2:1. For the MMWHS dataset, we split the training and test data to include individual patients only (no paired data across training and testing).}
                \label{fig:data_split_dgtta}
            \end{figure*}

            \subsection{Datasets}
                \subsubsection{BTCV: Multi-Atlas Labeling Beyond the Cranial Vault}
                    \label{sec:BTCV_dataset}
                    The dataset \citep{landman2015miccai} contains 30 labeled abdominal CT scans of a colorectal cancer chemotherapy trial with 14  organs: Spleen (SPL), right kidney (RKN), left kidney (LKN), gallbladder (GAL), esophagus (ESO), liver (LIV), stomach (STO), aorta (AO), inferior vena cava (IVC), portal vein and splenic vein (PSV), pancreas (PAN), right adrenal gland (RAG) and left adrenal gland (LAG).
                    Data dimensions reach from \sivoxtuple{512}{512}{85} to \sivoxtuple{512}{512}{198} and fields of view from \simmtuple{280}{280}{280} to \simmtuple{500}{500}{650}. We split the dataset into a 20/10 training/test set for our experiments and used a subset of ten classes that are uniformly labeled in all scans.

                \subsubsection{AMOS: A Large-Scale Abdominal Multi-Organ Benchmark
                for Versatile Medical Image Segmentation}
                    \label{sec:amos_dataset}
                    The AMOS dataset \citep{ji2022amos} consists of CT and MRI scans from eight scanners with a similar field of view as the BTCV dataset of patients with structural abnormalities in the abdominal region (tumors, etc.). Unlike the BTCV dataset's organs, AMOS has additional segmentation labels for the duodenum, bladder, and prostate/uterus but not for the PSV class.

                \subsubsection{MMWHS: Multi-Modality Whole Heart Segmentation}
                    \label{sec:MMWHS_dataset}
                    This dataset \citep{zhuang2019evaluation} contains CT and MR images of seven cardiac structures: Left ventricle, right ventricle, left atrium, right atrium, the myocardium of left ventricle, ascending aorta, and pulmonary artery.
                    The CT data resolution is \simmpervoxtuple{0.78}{0.78}{0.78}.
                    The cardiac MRI data was obtained from two sites with a 1.5 T scanner and reconstructed to obtain resolutions from \simmpervoxtuple{0.80}{0.80}{1.00} down to \simmpervoxtuple{1.00}{1.00}{1.60}.

                \subsubsection{SPINE: MyoSegmenTUM spine}
                    \label{sec:SPINE_dataset}
                    This MRI dataset \citep{burian2019lumbar} contains water, fat, and proton density-weighted lumbal spine scans with manually labeled vertebral bodies L1 --- L5.
                    The field of view spans \simmtuple{220}{220}{80} with a resolution of \simmpervoxtuple{1.8}{1.8}{4.0}.

                \subsubsection{TS: TotalSegmentator, 104 labels}
                    \label{sec:TS_dataset}
                    The large-scale TS dataset contains CT images of 1204 subjects with 104 annotated classes. The annotations were created semi-automated, where a clinician checked every annotation. The data was acquired across eight sites on 16 scanners with varying slice thickness and resolution  \citep{wasserthal2023totalsegmentator}.
                    Differing from the SPINE dataset, the vertebral bodies and the spinous processes are included in the class labels of this dataset, and intermediate model predictions were corrected accordingly in a postprocessing step to obtain reasonable results for evaluation.

                \subsubsection{Pre-/postprocessing}
                    We resampled all datasets to a uniform voxel size of \simmpervoxtuple{1.50}{1.50}{1.50}. For the SPINE task, we cropped the TS ground truth to omit the spinous processes with a mask dilated five voxels around the proposed prediction in the \fromto{TS}{SPINE} out-of-domain prediction setting to provide comparable annotations.

    \subsection{Related work}
        \label{sec:related_work}
        \paragraph{Domain generalization} 
            One way to improve model generalization is to increase the data manifold
            by augmentation.
            Augmentations can comprise simple intensity-based modifications such as the application of random noise, 
            partial corruption of image areas \citep{he2022masked,hoyer2023mic}, randomly initialized weights \citep{xu2020robust,ouyang2022causality} or differentiable augmentation schemes \citep{hu2022domain}.
            Generalization by domain randomization \citep{tobin2017domain} leverages a complete virtual simulation of input data to provide broadly varying data \citep{billot2023synthseg}.
            Using specialized self-supervised training routines has also proven to effectively improve model generalization \citep{bucci2021self, zhou2021models}.

        \paragraph{Test-time adaptation}
            Test-time adaptation (TTA) is performed in the target data domain and can be limited to a single target sample without access to source data. Tent is an often cited approach and adapts batch normalization layers of the network by minimizing the prediction entropy  \citep{wang2020tent}. Other works successfully introduced auxiliary tasks \citep{karani2021test} or  priors to steer the adaptation like AdaMI \citep{bateson2020source}.
            RSA uses edge-guided diffusion models to translate images from the source to the target domain and selects the best-synthesized edge-image candidate by the consistency of predictions \citep{zeng2024reliable}.
            Autoencoders capturing the feature statistics can reduce implausible target segmentation output like TTA-RMI \citep{karani2021test} or \citep{he2020self, he2021autoencoder}.
            Approaches nearest to our proposed method use consistency-self-supervision schemes in combination with sample augmentation but introduce further model complexity with Mean teacher or domain adversarial additions \citep{varsavsky2020test,perone2019unsupervised}. Many of the mentioned methods employ 2D models for image segmentation due to the memory requirements of the pipeline elements.

    \subsection{Proposed method}
        \label{sec:method}
\begin{figure*}
            \centerline{\includegraphics[width=\linewidth]{\dgttaPath/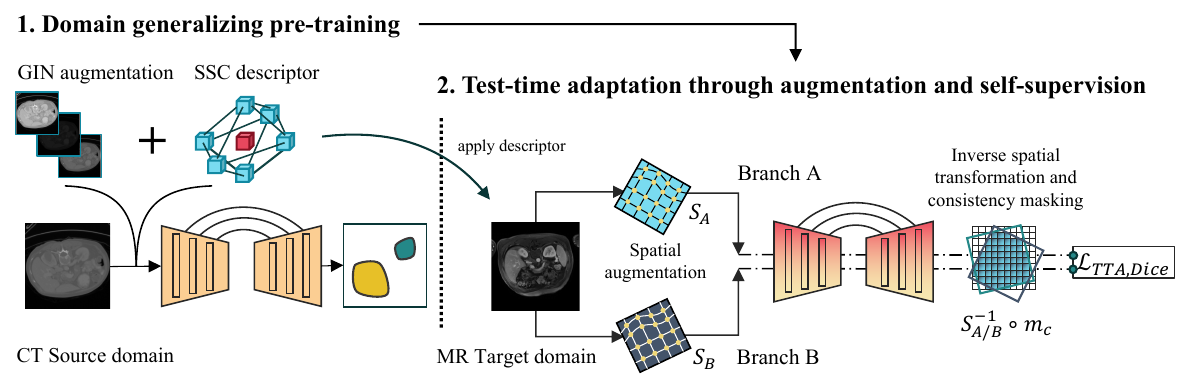}}
            \caption{Our proposed method consists of two steps that should be combined to reach optimal performance but can generally be used independently. Both steps rely on input feature modification to improve model generalization and enable unsupervised model adaptation at test time. Left: Model pre-training with source domain data. We propose to use GIN augmentation \citep{ouyang2022causality} and the SSC descriptor \citep{heinrich2013towards} in this step. Right: TTA is applied in the target data domain. Two different augmented versions of the same input are passed through the pre-trained segmentation network. The network weights are then optimized, supervising the predictions with a Dice loss and steering the network to produce consistent predictions. After inverse spatial transformations, consistency masking is applied to filter non-matching regions.}
            \label{fig:method_dgtta}
        \end{figure*}

        We seek to harness compact and effective domain-generalizing augmentation, as well as self-supervision during adaptation for 3D segmentation models to achieve optimal cross-domain performance.
        As shown in Fig. \ref{fig:method_dgtta} our method consists of two steps: Domain-generalized pre-training of the segmentation network involves using the domain-generalized techniques described below on the source image input. We use a cross-entropy and Dice loss objective \citep{isensee2021nnu}. Later, our TTA strategy is employed on individual target domain samples and does not require access to the source data. Both steps are integrated into the state-of-the-art nnUNet segmentation framework \citep{isensee2021nnu}.

        \subsection{Source domain domain-generalized pre-training}
            Pre-training is performed on the labeled source training dataset $D_{train}=\{\mathbf{x_s},\mathbf{y_s}\}_{s=1}^{l}$, $l\in\mathbb{N}$, where $\mathbf{x_s}$ and $\mathbf{y_s}$ can also be patches.
            Recently, global intensity non-linear augmentation GIN \citep{ouyang2022causality} was introduced to improve model generalization.
            In GIN, a shallow convolutional network $g$ is re-initialized at each iteration by random parameters $\rho$ and used to augment the input $\mathbf{x}$.
            The augmented image is then blended with the original image weighted by $\alpha$:
            \ifthenelse{\boolean{RENDER_TEXT_ONLY}}{}{
                \begin{equation}
                    \text{GIN}(\mathbf{x}) = \alpha\,g_\rho(\mathbf{x}) + (1-\alpha)\,\mathbf{x}
                \end{equation}
            }

            We propose combining GIN augmentation with self-similarity context (SSC) descriptors \citep{heinrich2013towards}. The approaches can be considered orthogonal, where GIN augmentation increases the input data manifold and SSC features were designed to yield one robust generalized description. Our intuition is that GIN-augmented features optimally enrich the SSC descriptor space and thus provide the network with meaningful input to generalize better.
            \ifthenelse{\boolean{RENDER_TEXT_ONLY}}{}{
                \begin{equation}
                    \text{SSC}(\mathbf{x,p,d}) = \text{exp}\left(-\frac{SSD(\mathbf{x}, \mathbf{p},\mathbf{d})}{\sigma^2_\mathcal{N}}\right),\quad \mathbf{p,d} \in \mathcal{N}, \quad \text{see \citep{heinrich2013towards}}
                \end{equation}
            }
            The generalizing SSC descriptor aggregates distance measures of the neighborhood around an image patch neglecting the image patch itself. For a given input image $\mathbf{x}$, smaller patches at location $\mathbf{p}$ are extracted and their feature distance to neighboring patches at a spatial distance $\mathbf{d}$ is evaluated. This difference is weighted by a local variance estimate $\sigma_\mathcal{N}$ \citep{heinrich2013towards}. The neighborhood pattern $\mathcal{N}$ diagonally connects adjacent patches $\mathbf{p}$ of the 6-neighborhood around the center patch resulting in a mapping of $\mathbb{R}^{1\times\lvert\Omega\rvert} \to \mathbb{R}^{12\times\lvert\Omega\rvert}$ for voxel space $\Omega$. For our experiments, we use a patch size and patch distance of \SI{1}{\vox}.

        \subsection{Target domain TTA}
            \label{sec:tta}
            Our test-time adaptation method can now be applied to pre-trained models: For any given pre-trained model $f_\theta$ on the training dataset $D_{train}$, we want to adjust the weights optimally to a single unseen sample of the target test set $D_{test}=\{\mathbf{x_t}\}$ during TTA.
            Instead of adding complex architectures, we propose to use two augmentation functions, $A$ and $B$, to obtain differently augmented images. The core idea of the method is to optimize the network to produce consistent predictions given two differently augmented inputs, where
            $S_{A/B}$ each denotes spatial augmentation:
            \ifthenelse{\boolean{RENDER_TEXT_ONLY}}{}{
                \begin{alignat}{2}
                    \mathbf{x_{A,t}} &= A(\mathbf{x_t}),\quad & A = S_A &\quad A: \mathbb{R}^{\lvert\Omega\rvert} \rightarrow \mathbb{R}^{\lvert\Omega\rvert} \\
                    \mathbf{x_{B,t}} &= B(\mathbf{x_t}),\quad & B = S_B &\quad B: \mathbb{R}^{\lvert\Omega\rvert} \rightarrow \mathbb{R}^{\lvert\Omega\rvert}
                \end{alignat}
            }
            Both augmented images are passed through the pre-trained network $f_\theta$:
            \ifthenelse{\boolean{RENDER_TEXT_ONLY}}{}{
                \begin{equation}
                    \mathbf{\hat{y}_{A/B,t}} = f_\theta(\mathbf{x_{A/B,t}})
                \end{equation}
            }
            Before calculating the consistency loss, both predictions $\mathbf{\hat{y}_{A/B,t}}$ need to be mapped back to the initial spatial orientation for voxel-wise compatibility by applying the inverse transformation operation $S_{A/B}^{-1}$. In addition, a consistency masking $m_{c}(\cdot)$ is applied to filter inversion artifacts with $\zeta$ indicating voxels that were introduced at the image borders during the inverse spatial transformation but are unrelated to the original image content:
            \ifthenelse{\boolean{RENDER_TEXT_ONLY}}{}{
                \begin{align}
                    m_c(\mathbf{\hat{y}_{A,t}}, \mathbf{\hat{y}_{B,t}}) &= \left[\mathbf{\hat{y}_{A,t}} \neq \zeta\right] \wedge \left[\mathbf{\hat{y}_{B,t}}  \neq \zeta\right]\\
                    A^{-1} &= m_c \circ S_{A}^{-1} \\
                    B^{-1} &= m_c \circ S_{B}^{-1}
                \end{align}
            }
            We steer the network to produce consistent outputs by comparing them after inversion and masking:
            \ifthenelse{\boolean{RENDER_TEXT_ONLY}}{}{
                \begin{equation}
                    \mathcal{L}_{TTA} = \ell(\mathbf{\hat{y}_{A,t}}, \mathbf{\hat{y}_{B,t}}) = \ell\left(A^{-1} \circ f_\theta\left(\mathbf{x_{A,t}}\right), B^{-1} \circ f_\theta\left(\mathbf{x_{B,t}}\right)\right)
                \end{equation}
            }
            As loss function $\ell$, we choose a Dice loss with predictions $\mathbf{\hat{y}_A}$ and $\mathbf{\hat{y}_B}$ given as class probabilities for all voxels in $\Omega$, where $e$ is a small constant ensuring numerical stability:
            \ifthenelse{\boolean{RENDER_TEXT_ONLY}}{}{
                \begin{equation}
                    \ell(\mathbf{\hat{y}_{A,t}}, \mathbf{\hat{y}_{B,t}}) = 1 - \frac{1}{\lvert \mathcal{B} \rvert \lvert \mathcal{C} \rvert}\sum_{}^{\mathcal{B},\mathcal{C}} \frac{\sum_{\omega}^{\Omega} 2 \cdot \hat{y}_{A,t,\omega} \cdot \hat{y}_{B,t,\omega} + e}{\sum_{\omega}^{\Omega}{\hat{y}_{A,t,\omega}}^d + {\hat{y}_{B,t,\omega}}^d + e}, \quad \omega \in \Omega
                    \label{eq:loss}
                \end{equation}
            }
            Selecting $d=2$ ensures consistency in the Dice loss landscape instead of $d=1$, which forces the network to additionally maximize the confidence of the prediction (see Fig. \ref{fig:loss_dgtta}).
            For spatial augmentation, we use affine image distortions on image/patch coordinates which we found to be sufficient during our experiments.

            \columnfigure{\dgttaPath/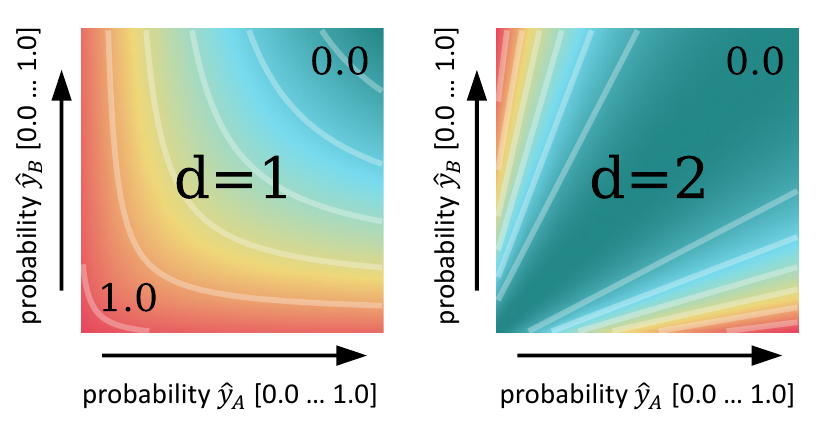}{1.}{Dice loss landscapes given scalar probability values $\hat{y}_A$ and $\hat{y}_B$ for different exponents $d=[1,2]$ in Eq. \ref{eq:loss}. $d=2$ yields zero loss along the diagonal, which is favorable for consistency.}{fig:loss_dgtta}
            
        \subsubsection{Optimization strategy}
            \label{sec:optimization_strategy}
            During TTA, only the classes $C$ of interest are optimized for consistency.
            To increase the robustness of predicted labels, we use an ensemble of three TTA models in the final inference routine of the nnUNet framework \citep{isensee2021nnu}.
            The AdamW optimizer was used with a learning rate of $\eta=1e{-5}$, weight decay $\beta=0.01$, and no scheduling. We empirically selected a count of $N_s = 12$ optimization steps throughout all of our experiments.
            Special caution has to be taken when applying test-time adaptation to models that require patch-based input. Since patch-based inference limits the field of view, the optimizer will adapt the model weights and overfit for consistency of the specific image region. Therefore, we accumulate gradients of $N_p = 16$ randomly drawn patches during one epoch step.

    \subsection{Statistical methods}
        In the following sections, segmentation quality is evaluated using the Dice score overlap metric (Dice) and the 95th percentile of the Hausdorff distance (HD95). The significance of TTA improvements is determined with the one-sided Wilcoxon Signed Rank test \citep{wilcoxon1992individual}, significance levels denoted as \textsuperscript{*}\mbox{p\textless0.05},  \textsuperscript{**}\mbox{p\textless0.01} and \textsuperscript{***}\mbox{p\textless0.001}) (software used: python 3.11, scipy 1.14.1). Tests were performed by C.W.

\section{Results}
    \subsection{Experiment I: Abdominal CT/MR cross-domain segmentation}
        \label{sec:exp1_results}
        In this experiment, we evaluate the performance of multiple base models and adapted models in an abdominal segmentation scenario. All base models were trained on source CT data (indicated by BS in the figures and tables). NNUNET denotes the standard model of the nnUNet pipeline \citep{isensee2021nnu} without specialized domain generalization capabilities. NNUNET BN denotes a nnUNet model with batch normalization layers. GIN, SSC and GIN+SSC base models were pre-trained with the domain generalizing techniques described in Sec. \ref{sec:method}. For comparison, we report the results of four related cross-domain methods as mentioned in Sec. \ref{sec:related_work} --- Tent, TTA-RMI, RSA and AdaMI \citep{wang2020tent,bateson2020source,karani2021test,zeng2024reliable}. Tent and AdaMI only need small changes to the pipeline (loss and layers) and we integrated them into the nnUNet pipeline. For the evaluation of TTA-RMI and RSA, we integrated the scenario data into the methods' pipelines. For AdaMI, a class ratio prior needs to be provided, which we estimated by averaging class voxel counts of the training dataset while we consider the same image field of view for the patch-based input.
        In addition to the base models' performances, we report the adapted models' performance after TTA (denoted by +A) and the significance of improvements compared to the NNUNET BS base model (reference). In the case of the batch normalization model NNUNET BN, we evaluated adapting only the normalization layer parameters (+A-nor) or the encoder (+A-enc) additionally to evaluate the adaptation of all parameters.
        An optimal configuration of our TTA routine was found in a preliminary combinatorial ablation study (see Appendix).

\begin{figure*}
        \centerline{\includegraphics[width=\linewidth]{\dgttaPath/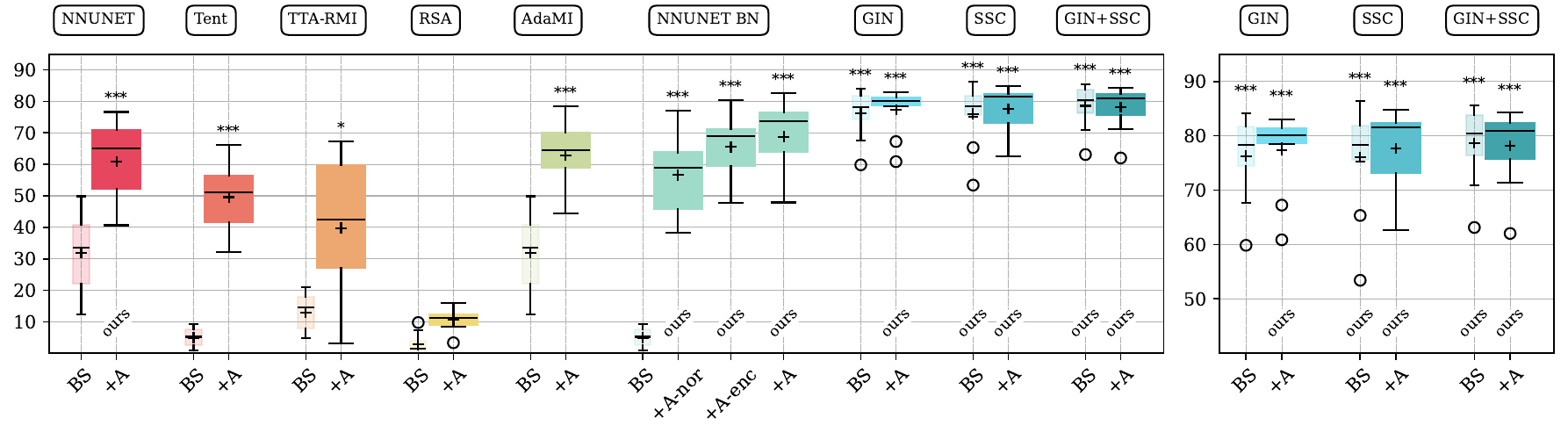}}
        \caption{Base (BS) and adapted model (+A) performance of several methods bridging a \fromto{CT}{MR} domain gap in abdominal organ segmentation. In the case of the batch normalization model NNUNET BN, we evaluated adapting only the normalization layer parameters (+A-nor) or the encoder (+A-enc) additionally to evaluate the adaptation of all parameters. \bplotdescription The right part of the figure shows a zoomed-in view of the three rightmost methods.}
        \label{fig:exp1_tta_dgtta}
    \end{figure*}

    Results can be compared via the boxplots in Fig. \ref{fig:exp1_tta_dgtta} or Tab. \ref{tab:exp1_dgtta} where Dice similarity is presented. Tab. \ref{tab:exp1_hd95_dgtta} presents Hausdorff distance results. For reference, we report the in-domain target data performance when training the NNUNET model on the target domain (no test data was included in target training).

    \begin{table*}
        \centering
        \caption{Base (BS) and adapted model (+A) performance given in Dice similarity \% of several methods bridging a \fromto{CT}{MR} domain gap in abdominal organ segmentation. In the case of the batch normalization model NNUNET BN, we evaluated adapting only the normalization layer parameters (+A-nor) or the encoder (+A-enc) additionally to evaluate the adaptation of all parameters. Higher Dice values indicate better performance. Mean column corresponds to values in Fig. \ref{fig:exp1_tta_dgtta}. Class names abbreviated: Spleen (SPL), right/left kidney (RKN/LKN), gallbladder (GAL), esophagus (ESO), liver (LIV), stomach (STO), aorta (AOR), inferior vena cava (IVC), and pancreas (PAN). The colors correspond to the label colors in Fig. \ref{fig:exp2_visual_dgtta}. Performance gains refer to the NNUNET BS model.}
        \label{tab:exp1_dgtta}
        \setlength{\tabcolsep}{3pt}
        \resizebox{\textwidth}{!}{
            \begin{tabular}{@{}llcS[table-figures-decimal=1,table-format=2.1]S[table-figures-decimal=1,table-format=2.1]S[table-figures-decimal=1,table-format=2.1]S[table-figures-decimal=1,table-format=2.1]S[table-figures-decimal=1,table-format=2.1]S[table-figures-decimal=1,table-format=2.1]S[table-figures-decimal=1,table-format=2.1]S[table-figures-decimal=1,table-format=2.1]S[table-figures-decimal=1,table-format=2.1]S[table-figures-decimal=1,table-format=2.1]S[table-figures-decimal=1,separate-uncertainty=true,table-format=2.1(3)]S[table-figures-decimal=1,table-format=-2.1,explicit-sign=+]@{}}
                \toprule
                \textbf{Method} & \textbf{Stage} & &
                \textbf{\legendbox{map10_10} SPL} &
                \textbf{\legendbox{map10_9} RKN} &
                \textbf{\legendbox{map10_8} LKN} &
                \textbf{\legendbox{map10_7} GAL} &
                \textbf{\legendbox{map10_6} ESO} &
                \textbf{\legendbox{map10_5} LIV} &
                \textbf{\legendbox{map10_4} STO} &
                \textbf{\legendbox{map10_3} AOR} &
                \textbf{\legendbox{map10_2} IVC} &
                \textbf{\legendbox{map10_1} PAN} &
                \textbf{Dice \boldmath{$\mu\pm\sigma$}} &
                \textbf{Gain} \\ \midrule
                \multirow{2}{*}{NNUNET} & \itshape BS & \itshape Reference & \itshape 40.2 & \itshape 21.9 & \itshape 15.9 & \itshape 24.9 & \itshape 22.9 & \itshape 76.0 & \itshape 34.3 & \itshape 26.4 & \itshape 21.8 & \itshape 35.3 & \itshape 32.0 \pm 16.3 & \\
                & +A & ours & 76.0 & 70.0 & 74.4 & 42.5 & 42.0 & 79.8 & 52.0 & 65.2 & 46.7 & 60.8 & 60.9 \pm 13.6 & +28.9 \\\hdashline\noalign{\vskip 0.5ex}
                \multirow{2}{*}{Tent} & BS & & 0.0 & 0.0 & 0.0 & 0.0 & 0.3 & 43.7 & 0.0 & 2.3 & 0.5 & 1.5 & 4.8 \pm 13.0 & -27.2 \\
                & +A & & 68.7 & 68.4 & 80.3 & 30.2 & 25.3 & 50.3 & 52.4 & 47.0 & 30.3 & 43.0 & 49.6 \pm 17.5 & +17.6 \\\hdashline\noalign{\vskip 0.5ex}
                \multirow{2}{*}{TTA-RMI} & BS & & 3.2 & 10.8 & 23.3 & 3.3 & 11.6 & 38.9 & 15.2 & 3.1 & 8.7 & 11.2 & 12.9 \pm 10.5 & -19.1 \\
                & +A & & 65.8 & 48.0 & 55.7 & 9.3 & 25.1 & 66.7 & 37.0 & 36.0 & 25.6 & 28.5 & 39.8 \pm 17.9 & +7.8 \\\hdashline\noalign{\vskip 0.5ex}
                \multirow{2}{*}{RSA} & BS & & 5.5 & 4.5 & 4.3 & 0.0 & 0.0 & 4.4 & 7.3 & 1.2 & 0.0 & 0.3 & 2.7 \pm 2.6 & -29.2 \\
                & +A & & 12.5 & 14.1 & 18.6 & 0.9 & 2.4 & 24.5 & 5.3 & 10.0 & 15.8 & 2.9 & 10.7 \pm 7.4 & -21.3 \\\hdashline\noalign{\vskip 0.5ex}
                \multirow{2}{*}{AdaMI} & BS & & 40.2 & 21.9 & 15.9 & 24.9 & 22.9 & 76.0 & 34.3 & 26.4 & 21.8 & 35.3 & 32.0 \pm 16.3 & \\
                & +A & & 75.0 & 77.7 & 83.1 & 37.1 & 37.9 & 83.8 & 56.5 & 68.1 & 48.2 & 61.1 & 62.8 \pm 16.7 & +30.9 \\\hdashline\noalign{\vskip 0.5ex}
                \multirow{4}{*}{NNUNET BN} & BS & & 0.0 & 0.0 & 0.0 & 0.0 & 0.3 & 43.7 & 0.0 & 2.3 & 0.5 & 1.5 & 4.8 \pm 13.0 & -27.2 \\
                & +A-nor & ours & 62.8 & 79.2 & 80.1 & 32.8 & 26.5 & 74.0 & 67.6 & 52.5 & 34.5 & 56.3 & 56.6 \pm 18.7 & +24.6 \\
                & +A-enc & ours & 80.1 & 87.1 & 88.4 & 34.6 & 30.9 & 83.1 & 73.9 & 68.8 & 43.8 & 64.8 & 65.5 \pm 20.5 & +33.6 \\
                & +A & ours & 81.6 & 87.2 & 89.2 & 40.1 & 35.9 & 84.7 & 73.4 & 75.1 & 54.1 & 66.8 & 68.8 \pm 18.3 & +36.8 \\\hdashline\noalign{\vskip 0.5ex}
                \multirow{2}{*}{GIN} & BS & & 81.9 & 90.4 & 91.9 & 63.7 & 47.8 & 92.7 & 73.2 & 80.9 & 72.1 & 68.1 & 76.3 \pm 13.5 & +44.3 \\
                & +A & ours & 81.6 & 90.5 & 92.1 & \B 72.3 & 48.9 & 93.3 & 74.7 & 79.2 & 70.9 & 70.6 & 77.4 \pm 12.6 & +45.4 \\\hdashline\noalign{\vskip 0.5ex}
                \multirow{2}{*}{SSC} & BS & ours & \B 83.6 & 92.9 & 93.0 & 60.5 & 39.1 & 93.1 & 74.6 & 81.1 & 69.5 & \B 73.2 & 76.1 \pm 16.1 & +44.1 \\
                & +A & ours & 83.4 & 91.2 & 92.9 & 68.1 & 48.6 & 91.4 & 74.4 & 83.3 & 71.3 & 72.2 & 77.7 \pm 13.0 & +45.7 \\\hdashline\noalign{\vskip 0.5ex}
                \multirow{2}{*}{GIN+SSC} & BS & ours & 83.0 & \B 93.3 & \B 93.3 & 65.9 & \B 50.3 & \B 94.1 & \B 76.5 & 83.9 & \B 73.7 & 71.9 & \B 78.6 \pm 13.3 & +46.6 \\
                & +A & ours & 82.2 & 92.7 & 92.7 & 68.4 & 47.1 & 93.4 & 74.5 & \B 84.8 & 73.5 & 72.5 & 78.2 \pm 13.6 & +46.2 \\\hdashline\noalign{\vskip 0.5ex}
                (NNUNET) & \multicolumn{2}{l}{(Target training)} & 86.5 & 94.6 & 95.0 & 70.9 & 59.9 & 97.3 & 81.4 & 91.2 & 85.4 & 83.2 & 84.5 \pm 11.1 & +52.6 \\
                \bottomrule
            \end{tabular}
        }
    \end{table*}
    \begin{table*}
        \centering
        \caption{Base and adapted model performance  given in the 95th percentile of the Hausdorff distance in mm (HD95) of several methods bridging a \fromto{CT}{MR} domain gap in abdominal organ segmentation. Smaller distances indicate better performance. Class names abbreviated: Spleen (SPL), right/left kidney (RKN/LKN), gallbladder (GAL), esophagus (ESO), liver (LIV), stomach (STO), aorta (AOR), inferior vena cava (IVC), and pancreas (PAN). The colors correspond to the label colors in Fig. \ref{fig:exp2_visual_dgtta}.  Distance reduction refers to the NNUNET BS model (the more negative, the better).}
        \label{tab:exp1_hd95_dgtta}
        \setlength{\tabcolsep}{3pt}
        \resizebox{\textwidth}{!}{
            \begin{tabular}{@{}llcS[table-figures-decimal=1,table-format=3.1]S[table-figures-decimal=1,table-format=3.1]S[table-figures-decimal=1,table-format=3.1]S[table-figures-decimal=1,table-format=3.1]S[table-figures-decimal=1,table-format=3.1]S[table-figures-decimal=1,table-format=3.1]S[table-figures-decimal=1,table-format=3.1]S[table-figures-decimal=1,table-format=3.1]S[table-figures-decimal=1,table-format=3.1]S[table-figures-decimal=1,table-format=3.1]S[table-figures-decimal=1,separate-uncertainty=true,table-format=3.1(3)]S[table-figures-decimal=1,table-format=-2.1,explicit-sign=+]@{}}
                \toprule
                \textbf{Method} & \textbf{Stage} & &
                \textbf{\legendbox{map10_10} SPL} &
                \textbf{\legendbox{map10_9} RKN} &
                \textbf{\legendbox{map10_8} LKN} &
                \textbf{\legendbox{map10_7} GAL} &
                \textbf{\legendbox{map10_6} ESO} &
                \textbf{\legendbox{map10_5} LIV} &
                \textbf{\legendbox{map10_4} STO} &
                \textbf{\legendbox{map10_3} AOR} &
                \textbf{\legendbox{map10_2} IVC} &
                \textbf{\legendbox{map10_1} PAN} &
                \textbf{HD95 \boldmath{$\mu\pm\sigma$}}  &
                \textbf{Reduction} \\ \midrule
                \multirow{2}{*}{NNUNET} & \itshape BS & \itshape Reference & \itshape 96.2 & \itshape 60.2 & \itshape 105.0 & \itshape 66.0 & \itshape 68.8 & \itshape 151.3 & \itshape 181.1 & \itshape 127.7 & \itshape 115.2 & \itshape 72.6 & \itshape 104.4 \pm 38.1 & \\
                & +A & ours & 70.9 & 36.5 & 58.2 & 93.8 & 51.5 & 158.1 & 191.4 & 126.3 & 116.5 & 72.8 & 97.6 \pm 47.3 & -6.8 \\\hdashline\noalign{\vskip 0.5ex}
            \multirow{2}{*}{Tent} & BS & & {\textemdash} & 90.1 & 102.7 & 32.7 & 141.5 & 46.0 & 93.5 & 182.5 & 123.0 & 76.3 & 98.7 \pm 43.7 & -5.7 \\
                & +A & & 186.7 & 215.1 & 181.7 & 130.9 & 180.3 & 239.3 & 228.3 & 171.4 & 162.8 & 182.9 & 187.9 \pm 30.5 & +83.5 \\\hdashline\noalign{\vskip 0.5ex}
            \multirow{2}{*}{TTA-RMI} & BS & & 105.9 & 101.0 & 92.6 & 69.4 & 149.9 & 114.3 & 163.3 & 101.9 & 109.6 & 96.8 & 110.5 \pm 26.0 & +6.1 \\
                & +A & & 67.9 & 47.6 & 84.8 & 65.9 & 122.0 & 93.5 & 103.4 & 99.7 & 112.9 & 76.9 & 87.4 \pm 22.0 & -17.0 \\\hdashline\noalign{\vskip 0.5ex}
            \multirow{2}{*}{RSA} & BS & & 116.8 & 88.9 & 60.2 & 77.7 & 129.7 & 206.9 & 115.7 & 146.1 & 154.1 & 101.3 & 119.7 \pm 40.2 & +15.3 \\
                & +A & & 80.2 & 119.9 & 102.8 & 97.7 & 116.1 & 83.9 & 100.2 & 43.4 & 36.6 & 86.6 & 86.7 \pm 26.4 & -17.7 \\\hdashline\noalign{\vskip 0.5ex}
            \multirow{2}{*}{AdaMI} & BS & & 96.2 & 60.2 & 105.0 & 66.0 & 68.8 & 151.3 & 181.1 & 127.7 & 115.2 & 72.6 & 104.4 \pm 38.1 & \\
                & +A & & 39.3 & 46.1 & 44.7 & 85.5 & 40.9 & 174.4 & 196.1 & 106.2 & 97.1 & 56.2 & 88.7 \pm 53.7 & -15.8 \\\hdashline\noalign{\vskip 0.5ex}
            \multirow{4}{*}{NNUNET BN} & BS & & {\textemdash} & 90.1 & 102.7 & 32.7 & 141.5 & 46.0 & 93.5 & 182.5 & 123.0 & 76.3 & 98.7 \pm 43.7 & -5.7 \\
                & +A-nor & ours & 50.9 & 26.5 & 37.7 & 52.4 & 71.2 & 180.7 & 87.9 & 110.5 & 50.3 & 35.2 & 70.3 \pm 44.1 & -34.1 \\
                & +A-enc & ours & 27.4 & 11.8 & 9.6 & 43.5 & 68.5 & 157.5 & 56.4 & 57.7 & 37.3 & 17.4 & 48.7 \pm 41.1 & -55.7 \\
                & +A & ours & 38.1 & 54.0 & 42.6 & 50.4 & 33.0 & 134.6 & 70.2 & 82.9 & 31.6 & 20.2 & 55.8 \pm 31.7 & -48.6 \\\hdashline\noalign{\vskip 0.5ex}
            \multirow{2}{*}{GIN} & BS & & \B 9.5 & 4.8 & 4.5 & 10.5 & 43.2 & 59.8 & 30.1 & 52.9 & 35.8 & 28.0 & 27.9 \pm 19.1 & -76.5 \\
                & +A & ours & 18.1 & 9.8 & 4.7 & 8.5 & 50.6 & 57.9 & 47.4 & 77.7 & 33.0 & 63.1 & 37.1 \pm 24.6 & -67.3 \\\hdashline\noalign{\vskip 0.5ex}
            \multirow{2}{*}{SSC} & BS & ours & 19.5 & 3.7 & 4.3 & 9.5 & 19.0 & \B 20.5 & 26.5 & 55.3 & 38.3 & 37.5 & 23.4 \pm 15.6 & -81.0 \\
                & +A & ours & 33.9 & 54.8 & 4.1 & 7.9 & 41.8 & 43.7 & 41.4 & 48.5 & 41.4 & 22.3 & 34.0 \pm 16.2 & -70.4 \\\hdashline\noalign{\vskip 0.5ex}
            \multirow{2}{*}{GIN+SSC} & BS & ours & 24.5 & \B 3.5 & \B 3.9 & 8.5 & \B 10.7 & 21.4 & \B 20.0 & 55.7 & 19.6 & 11.5 & 17.9 \pm 14.4 & -86.5 \\
                & +A & ours & 42.3 & 3.8 & 4.1 & \B 7.5 & 11.2 & 23.8 & 23.7 & \B 27.5 & \B 17.4 & \B 10.9 & \B 17.2 \pm 11.6 & -87.2 \\\hdashline\noalign{\vskip 1ex}
                (NNUNET) & \multicolumn{2}{l}{(Target training)} & 2.9 & 6.2 & 7.7 & 6.2 & 7.0 & 18.8 & 19.2 & 6.3 & 5.3 & 8.4 & 8.8 \pm 5.3 & -95.6 \\
                \bottomrule
            \end{tabular}
        }
    \end{table*}

        The NNUNET base model achieves a mean Dice value of \nnunetBaseVal{} when predicting across domains. This is a drop of \targetDrop{} compared to the NNUNET model when trained in the target data domain (reference model).
        Tent can outperform the reference model by \tentGain{} Dice with significant improvements. Applying TTA-RMI outperforms the reference model by \ttaRmiGain{} Dice.
        With RSA we experience a decrease to \rsaAfterAdap{} Dice. The mean performance before adaptation is \rsaBeforeAdaptation{} Dice.
        AdaMI performs best among the comparison methods with a mean Dice value of \adamiafteradaptation{} after adaptation.

        Generalizing pre-trained models achieve \ginBcvAmosVal, \sscBtcvAmosVal, \ginSscBtcvAmosVal{} Dice when predicting across domains (GIN, SSC, GIN+SSC). Subsequent adaptation gains \ginBtcvAmosAdapGain, \sscBtcvAmosAdapGain, \ginSscBtcvAmosAdapGain{} Dice. The highest mean performance is reached by the GIN+SSC pre-trained model (\ginSscBtcvAmosVal) whereas its adaptation results in a slight performance decrease (\ginSscBtcvAmosAdapGain). HD95 distance can be reduced to a lowest of \ginSscAdapBtcvAmosHdVal{} after adaptation of the GIN+SSC model.
        Using our TTA scheme, the performance of non-generalizing pre-trained models improves significantly. The highest model internal improvement is reached for the NNUNET BN model when all model parameters are adapted during TTA (\nnunetBnInternalGainBtcvAmosDice{} Dice and \nnunetBnInternalReductionBtcvAmosHd{} HD95). Adapting partial layers of the model results in lower gains.

        The results are reflected by the Hausdorff distance measurements (HD95) in Tab. \ref{tab:exp1_hd95_dgtta}. For Tent an increase of mean HD95 distance is measured whereas Dice performance increased. This can be explained by falsely predicted pixels, that appeared at the outside regions of the image far away from the organs' centers.

    \subsection{Experiment II: Multi-scenario CT/MR cross-domain segmentation with DG-TTA}
        \label{sec:exp2_results}
        Building upon Experiment I, we show the efficacy of our method leveraging the TS dataset 600 training samples as a strong basis in three segmentation tasks (all \fromto{CT}{MR}): Abdominal organ-, lumbar spine- and whole-heart segmentation (\fromto{TS}{SPINE}, \fromto{TS}{AMOS}, \fromto{TS}{MMWHS MR}).
        Opposed to the large TS dataset, we present results for the whole-heart segmentation task using only as few as 12 CT samples in model pre-training (\fromto{MMWHS CT}{MR}).

\begin{figure*}
            \centering
            \centerline{\includegraphics[width=\linewidth]{\dgttaPath/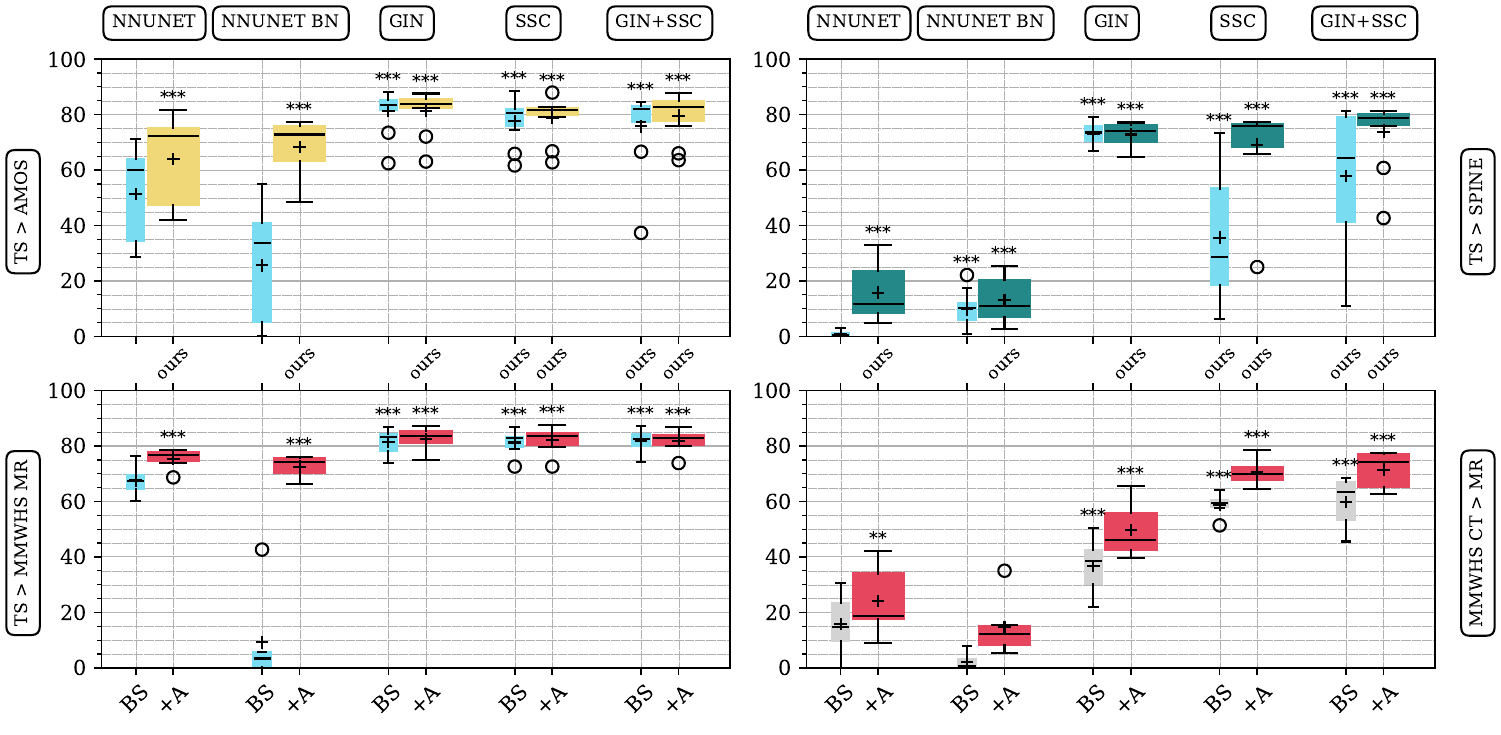}}
            \caption{Base (BS) and adapted (+A) model performance given in Dice similarity \%  for several cross-domain prediction scenarios. Top row and bottom left: TS pre-trained models. Bottom right: MMWHS CT pre-trained models with only 12 training samples. \bplotdescription}
            \label{fig:exp2_all_dgtta}
        \end{figure*}

\begin{figure*}
            \centering
            \centerline{\includegraphics[width=\linewidth]{\dgttaPath/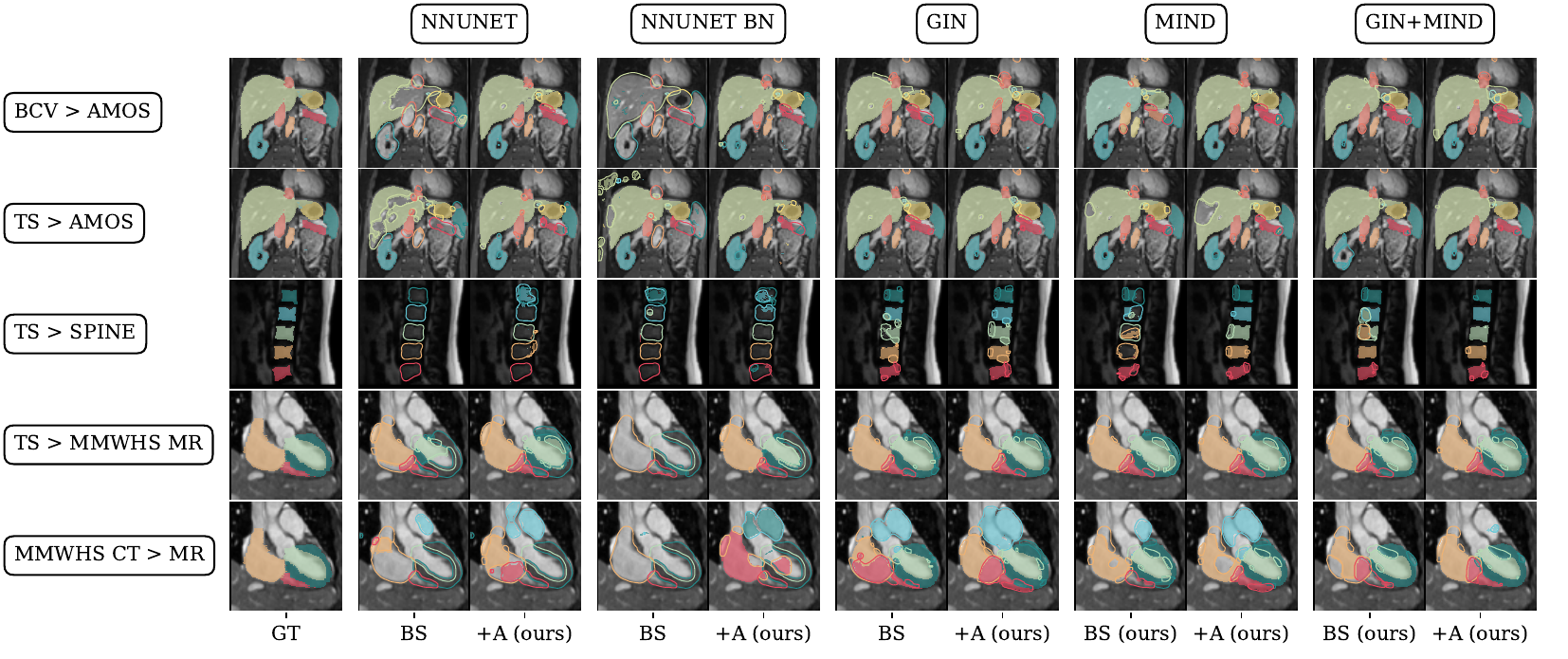}}
            \caption{Visual results correspond to statistics of Fig. \ref{fig:exp1_tta_dgtta} and \ref{fig:exp2_all_dgtta}. The rows show source and target datasets used; columns indicate the base (BS) or adapted (A+) models' prediction. Ground truth is given in the leftmost column. Positively predicted voxels are shown in colors. The erroneous area of predictions is marked with contours. The class colors for the abdominal task can be found in Tab. \ref{tab:exp1_dgtta}. Whole-heart class labels comprise the right ventricle \legendbox{map5_1}, right atrium \legendbox{map5_2}, left ventricle \legendbox{map5_3}, left atrium \legendbox{map5_4}, and myocardium \legendbox{map5_5}. Best viewed digitally.}
            \label{fig:exp2_visual_dgtta}
        \end{figure*}

        Abdominal prediction across domains using the \fromto{TS}{AMOS} datasets resulted in \nnunetAdapTsAmosVal, \nnunetBnAdapTsAmosVal, \ginAdapTsAmosVal, \sscAdapTsAmosVal, and \ginSscAdapTsAmosVal{} Dice similarity after adaptation of the models (see Fig. \ref{fig:exp2_all_dgtta}). All adapted and generalized pre-trained models could significantly increase the base model's performance. Compared to the more limited BTCV training dataset tested in experiment I (GIN+SSC+A, \ginSscBtcvAmosVal Dice), training on the TS dataset led to better top results (GIN+A, \ginAdapTsAmosVal{} Dice).
        The best mean Dice score for lumbar spine segmentation was reached by the GIN+SSC adapted model (\ginSscTsSpineAdap{} Dice).
        In the cardiac segmentation scenario rich- and low-sample training datasets were compared. For cross-domain prediction with the rich-sample pre-trained TS model a best mean Dice of \ginTsMmwhsMRVal{} was reached with GIN augmentation. For the low-sample training MMWHS dataset, GIN+SSC+A reached the highest mean Dice of \ginSscMmwhsCtMrAdapVal. Visual results of the mentioned scenarios are depicted in Fig. \ref{fig:exp2_visual_dgtta}.

        Tab. \ref{tab:exp2_score_ranks_dgtta} summarizes the mean Dice and HD95 scores of all scenarios for GIN, SSC and GIN+SSC methods along with their ranked scores.

        \begin{table}
            \centering
            \caption{Mean base (BS) and adapted (A+) model performance for GIN, SSC and GIN+SSC methods summarized for all evaluated scenarios. Performance given in Dice in \% and the 95th percentile of the Hausdorff distance (HD95) in mm. Rank of the scores per group is given in brackets. The combined score rank is given in the last column of the lower table group (mean of ranks across all scores per method).}
            \label{tab:exp2_score_ranks_dgtta}
            \resizebox{\textwidth}{!}{%
                \begin{tabular}{@{}llccccccclccccccc@{}}
                    \toprule
                    \textbf{Method} & \textbf{Stage} & & \textbf{\thead{TS \textgreater{} AMOS\\ Dice}} & \textbf{\thead{TS \textgreater{} AMOS\\ HD95}} & \textbf{\thead{TS \textgreater{} SPINE\\ Dice}} & \textbf{\thead{TS \textgreater{}SPINE\\ HD95}} & \textbf{\thead{TS \textgreater{} MMWHS MR\\ Dice}} & \textbf{\thead{TS \textgreater{} MMWHS MR\\ HD95}} \\ \midrule
                    NNUNET & BS & \itshape{Reference} & \itshape{51.4}\hphantom{ (0)} & \itshape{68.2}\hphantom{ (0)} & \hphantom{0}\itshape{0.8}\hphantom{ (0)} & \itshape{59.2}\hphantom{ (0)} & \itshape{67.6}\hphantom{ (0)} & \itshape{34.8}\hphantom{ (0)} \\\hdashline\noalign{\vskip 1.0ex}
                    \multirow{2}{*}{GIN} & BS & & 81.2 (2) & \B 11.8 (1) & 73.2 (2) & 11.7 (3) & 81.4 (5) & 11.1 (6) \\
                    & +A & ours & \B 81.4 (1) & 14.2 (3) & 72.8 (3) & \hphantom{0}9.7 (2) & \B 82.6 (1) & 10.2 (5) \\\hdashline\noalign{\vskip 0.5ex}
                    \multirow{2}{*}{SSC} & BS & ours & 77.7 (5) & 16.7 (5) & 35.6 (6) & 22.8 (6) & 81.3 (6) & \hphantom{0}8.9  (2) \\
                    & +A & ours & 79.0 (4) & 22.6 (6) & 69.1 (4) & 13.3 (4) & 82.0 (2) & \B \hphantom{0}8.5  (1) \\\hdashline\noalign{\vskip 0.5ex}
                    \multirow{2}{*}{GIN+SSC} & BS & ours & 75.7 (6) & 14.8 (4) & 57.9 (5) & 15.3 (5) & 81.8 (4) & 10.1 (4) \\
                    & +A & ours & 79.6 (3) & 12.1 (2) & \B 73.7 (1) & \B \hphantom{0}9.4 (1) & 81.8 (4) & \hphantom{0}9.4  (3) \\\bottomrule
                    & & & & & & & \\
                    & & & & & & & \\
                    \toprule
                    \textbf{Method} & \textbf{Stage} & & \textbf{\thead{BTCV \textgreater{} AMOS\\Dice}} & \textbf{\thead{BTCV \textgreater{} AMOS\\HD95}} & \textbf{\thead{MMWHS CT \textgreater{} MR\\ Dice}} & \textbf{\thead{MMWHS CT \textgreater{} MR\\ HD95}} & & \textbf{\textsc{\thead{Combined\\score rank}}} \\\midrule
                    NNUNET & BS & \itshape{Reference} & \itshape{32.0}\hphantom{ (0)} & \itshape{104.4}\hphantom{ (0)} & \itshape{15.8}\hphantom{ (0)} & \itshape{147.8}\hphantom{ (0)} & & \\\hdashline\noalign{\vskip 1.0ex}
                    \multirow{2}{*}{GIN} & BS & & 76.3 (5) & 27.9 (4) & 36.8 (6) & 85.6 (6) & & 4.0 \\
                    & +A & ours & 77.4 (4) & 37.1 (6) & 49.7 (5) & 77.5 (5) & & 3.5 \\\hdashline\noalign{\vskip 0.5ex}
                    \multirow{2}{*}{SSC} & BS & ours & 76.1 (6) & 23.4 (3) & 58.8 (4) & 53.0 (4) & & 4.7 \\
                    & +A & ours & 77.7 (3) & 34.0 (5) & 70.5 (2) & 25.4 (2) & & 3.3 \\\hdashline\noalign{\vskip 0.5ex}
                    \multirow{2}{*}{GIN+SSC} & BS & ours & \B 78.6 (1) & 17.9 (2) & 59.9 (3) & 47.9 (3) & & 3.7 \\
                    & +A & ours & 78.2 (2) & \B 17.2 (1) & \B 71.5 (1) & \B 17.8 (1) & & \B 1.9 \\ \bottomrule
                \end{tabular}%
            }
        \end{table}

        We found GIN+SSC+A to perform best across all scenarios reaching a score rank of \ginSscAdapRank{} outperforming SSC+A and GIN+A (ranks \sscAdapRank{} and \ginAdapRank{}).

\section{Discussion}
    In this study, we examined the use of generalizing augmentation combined with a generalizing feature descriptor for cross-domain medical image segmentation. We could show that the GIN+SSC augmentation-descriptor combination is highly effective, especially with limited data samples in pre-training in our two-step approach.

    If the domain-generalized pre-training does not result in sufficient target domain performance, our test-time adaptation scheme recovers weakly performing networks. This is crucial as it is impossible to know the target domain properties a priori.
    Significant gains were achieved in all scenarios with the GIN+SSC combination especially in the cardiac \fromto{MMWHS CT}{MR} scenario where we report improvements from \ginSscVsGinOnlyGainMmwhsCtMr{} Dice for GIN+SSC over using GIN-only augmentation.
    The presented results are consistent across five challenging \fromto{CT}{MR} out-of-domain prediction scenarios spanning abdominal, cardiac, and spine segmentation with small- and large-scale datasets from 12 to 600 training samples.
    Crucially, our approach can address many performant public segmentation models that have been trained on large private and unshareable datasets without the need to control their training regime for domain generalization directly. This is, moreover, beneficial because model providers would usually opt for a less generalizable model if this led to higher in-domain performance. Here, our method reaches improvements of up to \nnunetBnInternalGainBtcvAmosDice{} Dice for non-domain-generalized pre-trained models (see Sec. \ref{sec:exp1_results}, model NNUNET BN).


    \textbf{Pertinent findings in this study:}
    The proposed GIN+SSC augmentation-descriptor scheme
    outperforms augmentation-only and descriptor-only  configurations in our pre-training and TTA pipeline with a best score rank of \ginSscAdapRank{}.
    Similar to earlier works, we updated the batch normalization layers of the models with our TTA scheme but mean performance is higher when using consistency loss vs. the entropy-based formulation of Tent. Also, adapting all parameters is preferred over just single layers or parts of the model with our consistency scheme.
    Many cross-domain methods require 2D models since the surrounding pipelines are complex and set limits to the base model memory size. Our compact scheme can be used with 3D models and does not require prior assumptions.

    \textbf{Differences with regard to existing literature:}
    TTA-RMI, AdaMI and RSA methods evaluated in this study are tailored to specific setups, datasets, and their properties.
    Adaptation with AdaMI was successful but requires a class-ratio prior. This assumption is especially hard to fulfill in patch-based frameworks, where it is unclear, what organ is visible and how large it will appear in the image region.
    We could not acquire high scores with RSA which generates convincing source-style CT images from MR inputs but the predicted 2D masks are scattered in 3D space since the method does not include 3D convolutional layers.
    In our experience, there is a tendency towards designing overly complex methods.
    Our method is compact and readily integrated into the well-established nnUNet pipeline \citep{isensee2021nnu} since solely input-feature modifications and self-supervised test-time adaptation needs to be added.

    \textbf{Limitations of the technical method:} GIN augmentation alone can reach sufficiently high-performance out-of-domain when considering anatomies for which large pre-training dataset are available. Here our TTA scheme yields only moderate additional gains (abdominal \ginBtcvAmosAdapGain{} and \ginTsAmosAdapGain{}, lumbar spine \ginTsSpineAdapGain{} and cardiac \ginTsMmwhsMrAdapGain{} Dice overlap).
    We empirically selected the number of adaptation epochs and witnessed further score gains or sometimes score drops for samples if the test-time adaptation continued further as the specified epoch. A clear measure of convergence to stop the adaptation would be needed but is currently out of reach (since TTA has no ground truth target to evaluate). We kept the chosen number of epochs throughout all performed experiments and scenarios and are thus confident that the choice is reasonable for new scenarios as well.
    Apart from intensity-level domain gaps, domain gaps in image orientation, and resolution of the target domain scans impose further difficulties for cross-domain predictions. To mitigate these issues, the pre-training of our source models could further be optimized by multi-resolution augmentation in this regard or images of the unseen target data could be reoriented to a standardized orientation.

    \textbf{Conclusion:}
    Our study examined the use of a generalizing augmentation-descriptor combination for cross-domain segmentation and results indicate that its usage in pre-training and during test-time adaptation enhances cross-domain CT to MR prediction.

    \section*{Acknowledgements}
        This work was supported by the German Federal Ministry of Education and Research (BMBF) under grant ``MDLMA'' (031L0202B) and grant ``Medic V-Tach'' (01KL2008), the latter within the European Research Area Network on Cardiovascular Diseases (ERA-CVD).
    \printbibliography

    
    
    
    

\end{document}